\ifx\pdfoutput\undefined
\else
  \pdfoutput=1
\fi
\documentclass[]{xiaomiev}

\usepackage[toc,page,header]{appendix}
\usepackage{minitoc}
\usepackage{solarized-light}
\usepackage{booktabs}
\usepackage{multirow}
\usepackage{graphicx}
\usepackage{array}
\usepackage{siunitx,array}
\usepackage[table]{xcolor}
\usepackage{makecell,array,booktabs}
\usepackage{makecell}
\usepackage{framed}
\usepackage{amsmath}
\usepackage{amssymb}
\usepackage{mathtools}
\usepackage{amsthm}
\usepackage{bbding}
\usepackage{hyperref}
\usepackage{amsmath} 
\usepackage{placeins}
\usepackage[colorinlistoftodos]{todonotes}
\usepackage{longtable}
\usepackage{hhline}
\usepackage{fancyvrb}
\usepackage{graphicx}
\usepackage[table]{xcolor}
\usepackage{booktabs}
\usepackage{float}
\usepackage{fvextra}
\usepackage{CJKutf8}
\usepackage{multicol}
\usepackage{float}
\usepackage{placeins}
\usepackage{cleveref}
\usepackage{tablefootnote}
\usepackage{threeparttable}
\usepackage{tabularx}
\usepackage{mdframed}
\usepackage{subcaption}
\usepackage[usestackEOL]{stackengine}
\usepackage[numbers]{natbib}
\newcommand{\commentout}[1]{}
\renewcommand{\paragraph}[1]{\noindent\textbf{#1.}\hspace*{1em}}

\usepackage{enumitem}
\usepackage{hyperref}
\setlist[itemize]{leftmargin=15pt}
\usepackage{siunitx,array}
\RequirePackage{xspace}
\makeatletter
\DeclareRobustCommand\onedot{\futurelet\@let@token\@onedot}
\def\@onedot{\ifx\@let@token.\else.\null\fi\xspace}

\def\ie{\emph{i.e}\onedot}

\makeatother

\title{Walk With Me: Long-Horizon Social Navigation for Human-Centric Outdoor Assistance}

\author{
  Lingfeng Zhang$^{1,2,3}$, 
  Xiaoshuai Hao$^{3,\dagger,\ddagger}$,
  Xizhou Bu$^{3,4}$,
  Yingbo Tang$^{5}$, \\
  Hongsheng Li$^{1}$,
  Jinghui Lu$^{3}$,
  Xiu-shen Wei$^{6}$,
  Jiayi Ma$^{7}$,
  Yu Liu$^{8}$,
   \\
   Jing Zhang$^{7}$,
  Hangjun Ye$^{3}$,
  Xiaojun Liang$^{2}$,
  Long Chen$^{3}$,
  Wenbo Ding$^{1,\dagger}$
}

\affiliation[1]{Tsinghua University}
\affiliation[2]{Pengcheng Laboratory}
\affiliation[3]{Xiaomi EV}
\affiliation[4]{Fudan University\\ $^{5}$Institute of Automation, Chinese Academy of Sciences}
\affiliation[6]{Southeast University\\ $^{7}$Wuhan University}
\affiliation[8]{Hefei University of Technology}

\contribution[\dagger]{Corresponding authors}
\contribution[\ddagger]{Project leader}

\website{\url{https://linglingxiansen.github.io/Walk-with-Me/}}

\abstract{
Assisting humans in open-world outdoor environments requires a robot to translate high-level natural-language intentions into safe, long-horizon, and socially compliant navigation behavior. This capability is essential for human-centric applications such as last-mile delivery and blind guidance, yet remains challenging in complex outdoor scenarios. Existing map-based methods typically rely on costly pre-built HD maps, which are difficult to construct and maintain at scale, while still offering limited semantic understanding of high-level human intent. In contrast, learning-based navigation policies are mostly restricted to indoor and short-horizon settings. To bridge this gap, we propose \textbf{\textit{Walk with Me}}, a map-free framework for long-horizon social navigation from high-level human instructions. Instead of requiring a pre-built HD map, Walk with Me only leverages GPS context and lightweight candidate points-of-interest from a public map API for semantic destination grounding and waypoint proposal. Specifically, a High-Level Vision-Language Model first grounds the user’s abstract instruction into a concrete destination and plans a coarse waypoint sequence. During execution, an observation-aware routing mechanism decides whether the current situation can be handled by the Low-Level Vision-Language-Action policy or should be escalated back to the High-Level VLM for explicit safety reasoning. Routine navigation segments are executed by the Low-Level VLA, which predicts socially compliant navigation actions from visual observations, the next waypoint, and recent trajectory context, while complex situations such as crowded crossings trigger high-level reasoning and stop-and-wait behavior when unsafe. The robot then executes the predicted actions, updates its state, and iterates until reaching the destination. By combining semantic intent grounding, map-free long-horizon planning, safety-aware high-level reasoning, and low-level action generation, \textbf{\textit{Walk with Me}} enables practical outdoor social navigation for human-centric assistance.}

\begin{document}
\maketitle
\vspace{-4pt}

\begin{figure}[!t]
    \centering
\includegraphics[width=\linewidth]{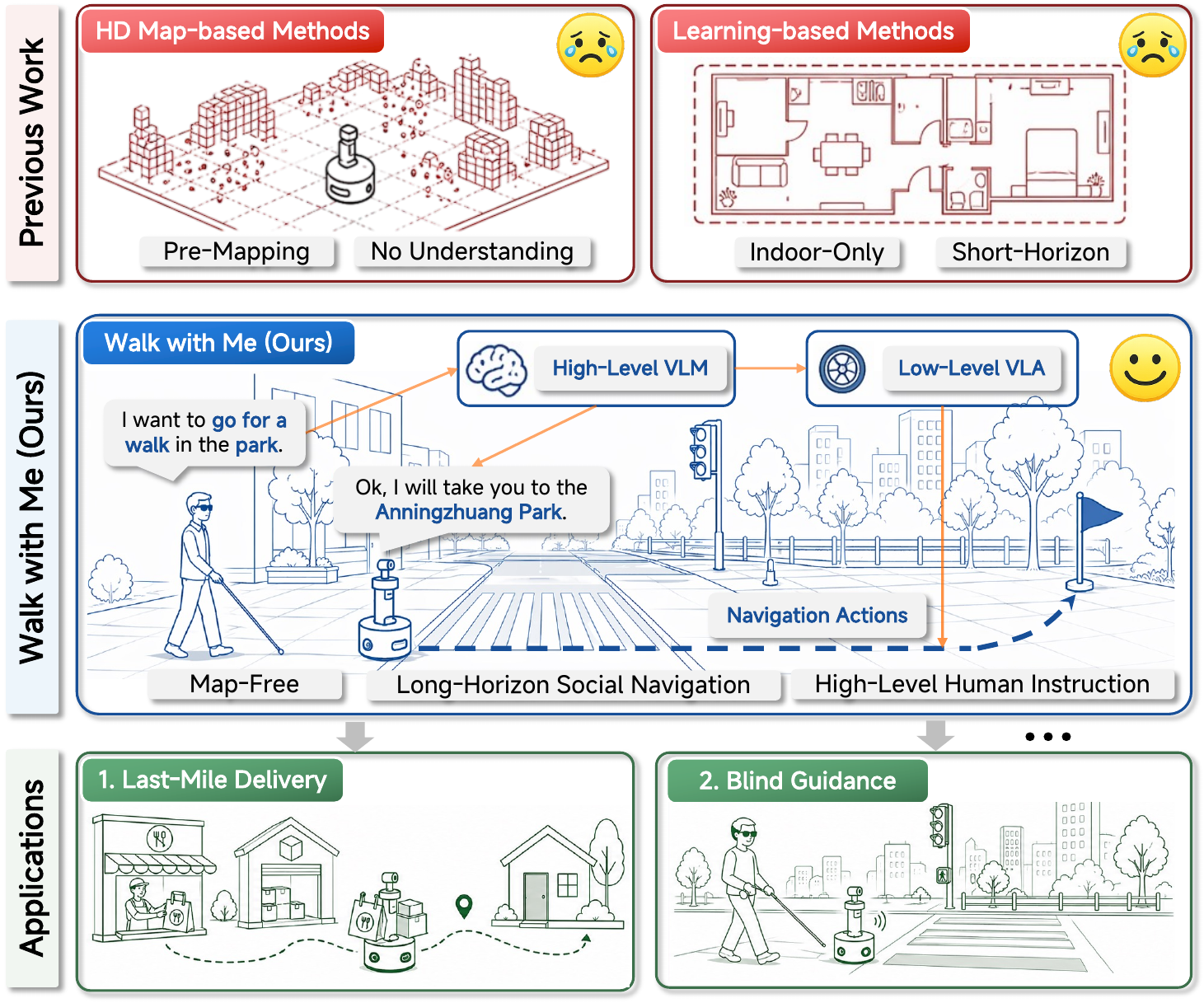}
    \caption{\textbf{Motivation of \textit{Walk with Me}.} Previous navigation methods typically rely on pre-built HD maps, manually specified goals, or short-horizon indoor settings, making them difficult to apply to open-world human-centric assistance. In contrast, our framework grounds high-level human instructions into real-world destinations, plans long-horizon routes with lightweight public map priors, and performs socially compliant navigation with explicit safety reasoning. This enables practical outdoor applications such as last-mile delivery and blind guidance.}
    \label{fig1}
\end{figure}


\section{Introduction}

As embodied agents move from controlled laboratories into everyday human environments, they are increasingly expected not only to navigate to predefined coordinates, but also to \emph{walk with humans} toward destinations described in natural language. Human-centric applications such as last-mile delivery and blind guidance require a robot to understand an abstract instruction such as ``I want to go for a walk'', infer a meaningful real-world destination, and accompany the user there in a safe and socially compliant manner~\cite{karnan2022socially,francis2023principles,zhang2026you,zhang2025nava,zhang2026embodiedplan,lu2026onevl}. We refer to this setting as \textit{outdoor social navigation from high-level human instructions}. Unlike conventional point-goal navigation or short-horizon indoor navigation, this problem simultaneously requires open-vocabulary intent grounding, long-horizon route reasoning, and socially aware local decision-making in dynamic outdoor scenes involving crosswalks, traffic, and crowds.

Existing navigation systems are not well suited to this setting. Classical map-based pipelines~\cite{thrun2002probabilistic,macenski2020marathon2,cadena2017past,fu2026sef} typically depend on pre-built high-definition (HD) maps that provide accurate metric geometry for localization and planning. Although such maps are effective in structured domains, they are costly to construct, difficult to maintain under frequent urban changes, and hard to scale to the long-tail outdoor environments that a human-assistance robot may encounter. More importantly, HD maps mainly encode geometry rather than human intent: they do not explain where a user wants to go when given an abstract request such as ``take me somewhere to relax''. In contrast, recent learning-based navigation policies, including end-to-end visual navigation models~\cite{shah2023gnm,shah2023vint,sridhar2024nomad,zhang2024trihelper,zhang2025multi,liu2025toponav} and vision-language-action (VLA) models~\cite{black2025pi05,bu2025univla,wuevaluating,hao2025mimo}, have shown promising closed-loop behaviors, but most of them remain limited to indoor environments, short-horizon routes, or low-level goal specifications. Without explicit long-horizon grounding and safety reasoning, these policies struggle to scale to city-level outdoor navigation and may produce socially disruptive behaviors in complex pedestrian scenes~\cite{mavrogiannis2023core,chen2025socialnav,zhang2025socialnavmap,xiao2025team}.

This raises a central question: \textit{how can a robot convert high-level human intent into long-horizon, socially compliant outdoor navigation without relying on pre-built HD maps?} Our key observation is that the information needed for global intent grounding and coarse route planning is already available in public map services. Such services provide lightweight geographic priors, including GPS locations, points of interest (POIs), and walking-route APIs, through a standard interface. This is fundamentally different from requiring the robot or the system operator to build and maintain an HD metric map. In this work, we use \textit{map-free} to mean that the robot does not depend on a pre-built HD map for navigation. Instead, it uses public map services only as a lightweight semantic and geographic prior for destination grounding and coarse waypoint proposal, while relying on onboard perception and learned policies for local navigation.

Based on this observation, we propose \textbf{\textit{Walk with Me}}, a hierarchical framework for map-free, long-horizon outdoor social navigation from high-level human instructions. The core idea is to decouple \emph{where to go} from \emph{how to walk there}. A \textbf{High-Level Vision-Language Model (VLM)} interprets the user's abstract instruction together with GPS context and candidate POIs from a public map service, selects a concrete destination, and generates a coarse sequence of geo-referenced waypoints. A \textbf{Low-Level Vision-Language-Action (VLA)} policy then takes the current visual observation, the next waypoint, and recent trajectory context to generate socially compliant local navigation actions. Between these two modules, an observation-aware router monitors the current scene and upcoming route segment. Routine segments are handled by the Low-Level VLA, while complex situations such as crosswalks, dense crowds, or ambiguous traffic conditions are escalated back to the High-Level VLM for explicit safety reasoning. When the scene is judged unsafe, the robot executes a stop-and-wait behavior until it is safe to proceed.

The overall pipeline forms a closed perception--reasoning--action loop. Given a user instruction, \textbf{\textit{Walk with Me}} first grounds the instruction into a destination and coarse route using GPS and public map-service candidates. During execution, the robot continuously updates its visual observation, pose, trajectory history, and next waypoint. The router decides whether the current step should be solved by low-level trajectory generation or high-level safety reasoning. The robot executes the resulting action, updates its state, and repeats this process until the destination is reached. This design allows the system to retain the scalability of public geographic services, the semantic reasoning ability of VLMs, and the closed-loop control capability of VLA policies, without requiring a pre-built HD map.

We deploy \textbf{\textit{Walk with Me}} on a commodity wheeled robot and evaluate it in diverse real-world outdoor environments. \textbf{\textit{Walk with Me}} successfully completes kilometer-scale navigation tasks from abstract human instructions and handles socially sensitive outdoor scenarios such as pedestrian crossings and crowded areas, demonstrating the usability of the proposed framework for practical human-centric outdoor assistance.

Our main contributions are summarized as follows:
\begin{itemize}
    \item We formulate \textit{map-free, long-horizon outdoor social navigation from high-level human instructions}, a human-centric navigation problem that requires intent grounding, coarse route reasoning, and socially compliant local control without relying on pre-built HD maps.
    \item We propose \textbf{\textit{Walk with Me}}, a hierarchical framework that couples a High-Level VLM for instruction grounding, destination selection, waypoint proposal, and safety reasoning with a Low-Level VLA for local socially compliant trajectory generation.
    \item We introduce an observation-aware routing mechanism that adaptively assigns routine navigation to the Low-Level VLA and escalates complex social situations, such as crossings and crowds, to the High-Level VLM for explicit safety reasoning and stop-and-wait decisions.
    \item We instantiate the framework as a real-world robotic system using public map services and a  wheeled robot, demonstrating reliable kilometer-scale outdoor social navigation from abstract human instructions. 
\end{itemize}
\section{Related Work}

\textbf{Social Navigation.}
Social navigation studies how robots move toward a goal in human-populated environments while not only avoiding collisions, but also respecting social norms such as personal space, passing conventions, motion legibility, and non-interference with human activities. Early work mainly addressed this problem through human-aware planning and hand-crafted cost design, where proxemic constraints, directional preferences, or socially weighted objectives were incorporated into classical planners. More recent research has increasingly shifted toward learning-based formulations, especially deep reinforcement learning, to capture the complex and highly interactive nature of human-robot navigation in crowded scenes. Representative methods such as Crowd-Robot Interaction~\cite{chen2019crowd}, DS-RNN~\cite{liu2021decentralized}, and SG-DQN~\cite{zhou2021sgdqn} learn socially aware local policies by modeling interactions between the robot and nearby pedestrians, while later works further improve foresight and social compliance by reasoning over richer interaction structures or human intentions, e.g., intention-aware crowd navigation~\cite{liu2023intention}. Another important line of work explicitly addresses uncertainty and partial observability in dynamic crowds. For example, PaS~\cite{mun2022pas} leverages social cues from visible pedestrians to infer occluded agents, Falcon~\cite{gong2024falcon} introduces future human trajectory prediction to discourage blocking behaviors, and SocialNav-Map~\cite{zhang2025socialnavmap} integrates predicted human motion into dynamic occupancy maps for zero-shot socially compliant navigation. These studies have substantially advanced short-horizon social navigation by improving safety, efficiency, and anticipatory behavior in crowded environments. However, most of them still assume a predefined geometric or point-goal destination and focus primarily on local decision making in indoor or spatially bounded settings, leaving open the problem of translating abstract human requests into long-horizon, socially compliant outdoor navigation~\cite{zhang2024trihelper,zhang2025multi,gong2025stairway,liu2025toponav,zhang2025nava,xiao2025team}.

\textbf{VLMs for Social Navigation.}
Recent progress in VLMs and VLAs has opened a new direction for social navigation by enabling robots to couple semantic understanding with action generation. In the broader embodied navigation and vision-and-language navigation literature, methods such as NavGPT~\cite{zhou2024navgpt}, MapGPT~\cite{chen2024mapgpt}, MapNav~\cite{zhang2025mapnav}, and NaVILA~\cite{cheng2024navila} demonstrate that large multimodal models can support instruction decomposition, landmark grounding, spatial memory construction, hierarchical planning, and action generation from egocentric observations~\cite{wuevaluating,hao2025mimo,zhang2025your,zhang2026you,zhang2025team,zhang2026embodiedplan}. While these methods are not designed specifically for social navigation, they highlight the potential of foundation models to connect high-level language understanding with embodied control~\cite{tang2025roboafford,hao2025roboafford++,hao2026h2r,zhang2025lips,zhang2025video,fu2026sef,li2026weather,zheng2025railway,cheng2025exploring}. More directly related to social navigation, recent works begin to use VLMs as social reasoning modules rather than pure instruction-following engines. VLM-Social-Nav~\cite{song2024vlmsocialnav} uses a VLM to score socially appropriate robot actions for a downstream planner, SocialNav-SUB~\cite{munje2025socialnavsub} systematically evaluates whether current VLMs can understand spatial, temporal, and social relations in robot navigation scenes, and SocialNav~\cite{chen2025socialnav} proposes a hierarchical reasoning-and-action framework that combines high-level social understanding with low-level trajectory generation. Follow-up works such as SocialNav-MoE~\cite{kawabata2025socialnavmoe}, E-SocialNav~\cite{xiao2026esocialnav}, and MAction-SocialNav~\cite{wang2025mactionsocialnav} further explore efficient small VLMs or LLMs, reinforcement fine-tuning, and ambiguity-aware decision making for socially compliant behavior. Together, these efforts suggest that foundation models are promising for bringing semantic and social reasoning into robot navigation. However, existing VLM-based social navigation methods remain largely centered on local scene understanding or short-horizon action selection, and generally do not address the full problem studied here: map-free, long-horizon outdoor assistance from high-level human instructions, which requires destination grounding, coarse route planning, and adaptive switching between low-level execution and explicit safety reasoning.
\section{Methodology}
\begin{figure*}[!t]
\centering
\includegraphics[width=\textwidth]{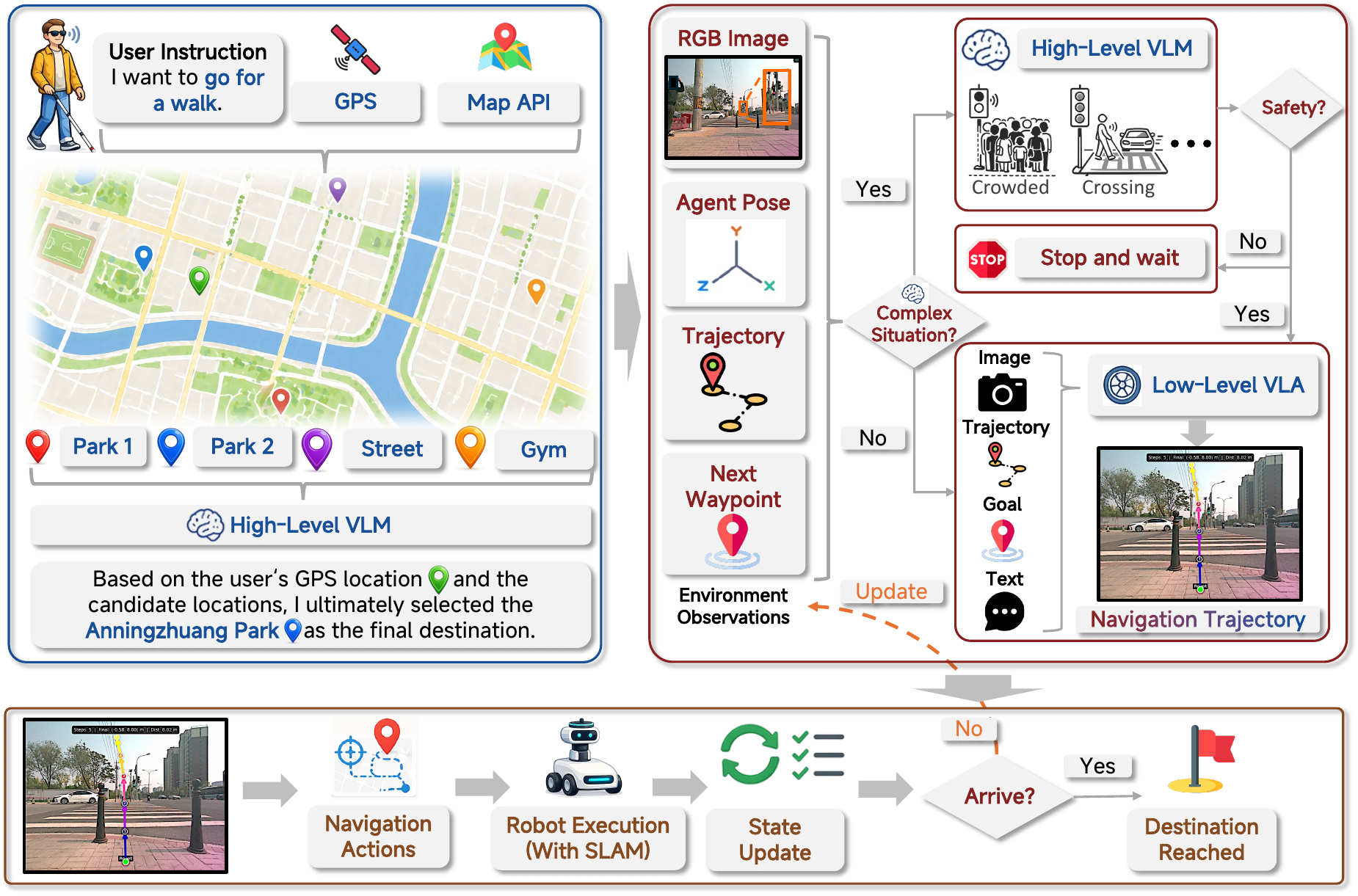}
\caption{
\textbf{Overall framework of \textit{Walk with Me}.}
\textbf{Left panel:} given an abstract human instruction (e.g., ``\textit{I want to go for a walk}''), a \textbf{High-Level VLM} reads the user's GPS context and a set of candidate POIs queried from a public map service, and selects a concrete destination with a natural-language rationale. A coarse walking route obtained from the map service's walking-route API is re-sampled into a sequence of geo-referenced waypoints.
\textbf{Right panel:} at every control step, the robot aggregates the current RGB image, agent pose, trajectory history, and next waypoint into an environment observation. The observation is sent to the \textbf{High-Level VLM}, which jointly assesses whether the current step is routine or complex and decides whether the robot should proceed or stop and wait. If the decision is to proceed, the step is dispatched to the \textbf{Low-Level VLA} for socially compliant local trajectory prediction; otherwise, the robot follows the High-Level VLM's stop-and-wait decision.
\textbf{Bottom panel:} the resulting trajectory or safety decision is executed by the robot, after which the state is updated and the loop repeats until the destination is reached.
}
\vspace{-1em}
\label{fig:framework}
\end{figure*}

\subsection{Problem Formulation and Framework Overview}

We consider a mobile robot that accompanies a human user to an open-world outdoor destination specified only by an abstract natural-language instruction $l$, such as ``I want to go for a walk.'' The robot is equipped with onboard visual perception, GPS context, and local state estimation for closed-loop execution. The only external geographic prior available to the system is a public map service, such as Baidu Maps, Google Maps, or Gaode Maps, which provides lightweight place and walking-route information.

The robot does not rely on a pre-built HD map. Instead, the public map service is used only as a lightweight semantic and geographic prior for destination grounding and coarse route construction. Local navigation, social interaction, and safety-aware behavior are handled by onboard perception and learned policies.

Given the high-level instruction $l$, the goal is to produce, at each control step $t$, a local navigation action $a_t$ such that the robot: 
(i) grounds the abstract instruction into a semantically appropriate destination;
(ii) follows a long-horizon route toward that destination;
(iii) generates socially compliant local motion; and
(iv) explicitly reasons about safety in complex outdoor scenarios such as crossings, dense crowds, and ambiguous traffic situations.

As shown in Fig.~\ref{fig:framework}, \textit{Walk with Me} addresses this problem with a hierarchical framework composed of three stages. First, a high-level semantic planner uses a Vision-Language Model (VLM), GPS context, and POI candidates from a public map service to select a concrete destination and construct a coarse waypoint sequence. Second, an adaptive dual-policy navigation module builds online observations and queries the High-Level VLM for a joint scene assessment and safety decision. The High-Level VLM determines whether the current step is routine or complex and simultaneously decides whether the robot should proceed or stop and wait. Steps with a proceed decision are dispatched to a Low-Level Vision-Language-Action (VLA) policy for local trajectory generation, while unsafe steps trigger stop-and-wait behavior. Third, a closed-loop execution module converts the selected trajectory or safety decision into robot actions, updates the robot state, and iterates until the destination is reached.

\subsection{High-Level Semantic Planning}

The high-level semantic planner converts an abstract user instruction into a concrete, robot-executable route. It first grounds the user intent into a destination using POI candidates from a public map service, and then transforms the resulting walking route into a sequence of coarse local waypoints.

\subsubsection{POI-aware Destination Grounding}

Given user instruction $l$ and coarse GPS context, the High-Level VLM first proposes a short list of semantic place categories that are consistent with the instruction. For example, for the instruction ``I want to go for a walk,'' the model may propose categories such as park, plaza, or riverside walkway.

Each category is then issued as a POI search query to the map service using the user's approximate location as geographic context. The map service returns a ranked candidate set
\begin{equation}
    C=\{c_i\},
\end{equation}
where each candidate $c_i$ contains lightweight place information such as a name, category, and approximate location. The High-Level VLM is prompted with the original instruction $l$, the current geographic context, and the candidate set $C$. It is asked to select a single concrete destination
$c^\star \in C$
and provide a short natural-language rationale grounded in the user intent and candidate attributes.

This design allows the robot to translate open-ended human intent into an explicit real-world goal without requiring a pre-built semantic map. The destination is selected from places available around the user, rather than from a closed set of manually predefined goals.

\subsubsection{Route Retrieval and Waypoint Construction}

Once the destination $c^\star$ is selected, the planner queries the walking-route API with the current location and the selected destination. The returned route provides a coarse pedestrian path and a sequence of turn-by-turn walking steps. These route steps may also contain lightweight semantic cues such as pedestrian-crossing or traffic-light information.

The route is then simplified into an ordered waypoint sequence used by the downstream navigation module. Each waypoint provides coarse directional guidance, and route-level semantic cues are preserved when available so that the robot can recognize segments that may require additional safety reasoning. Before entering the main navigation loop, the waypoint sequence is aligned with the robot's local execution frame, allowing downstream modules to operate on local navigation goals without requiring a pre-built HD map.

\subsection{Adaptive Dual-Policy Navigation}

After the high-level route has been constructed, the robot navigates in a closed loop. At each control step, it builds an environment observation and queries the High-Level VLM for a joint scene assessment and safety decision. The High-Level VLM determines whether the current situation is routine or safety-critical and simultaneously decides whether the robot should proceed or stop and wait. Based on this joint output, the system either invokes the Low-Level VLA for local trajectory generation or executes a stop-and-wait behavior.

At step $t$, the robot constructs an environment observation
\begin{equation}
    o_t=(I_t,s_t,H_t,w_t^{goal}),
\end{equation}
where $I_t$ is the current RGB image, $s_t$ is the current local pose estimate, $H_t$ is a short history of recent poses, and $w_t^{goal}$ is the current lookahead waypoint.

The lookahead waypoint is selected from the global waypoint sequence using a simple forward-looking rule. This produces a stable local navigation goal that advances smoothly along the route and naturally adapts to both straight paths and turns.

\subsubsection{Joint Routing and Safety Decision}

The observation-aware routing process is implemented through the High-Level VLM as a joint decision. Given the current visual observation and the upcoming route segment, the High-Level VLM determines whether the current control step corresponds to a routine navigation segment or a complex situation, while simultaneously deciding whether the robot should proceed or stop and wait.

Before querying the VLM, the system constructs a step-aware natural-language instruction $l_t$. This instruction is generated from the walking-route step associated with the current lookahead waypoint, for example indicating whether the robot should continue straight or prepare for a turn. This text grounds the joint decision and the local policy in the high-level route plan.

Formally, the High-Level VLM outputs
\begin{equation}
    \text{VLM}_{route}(I_t,l_t,w_t^{goal})
    \rightarrow
    (r_t,d_t,conf,reason),
\end{equation}
where
\begin{equation}
    r_t \in \{\text{routine}, \text{complex}\},
    \quad
    d_t \in \{\text{proceed}, \text{stop-and-wait}\}.
\end{equation}
The decision is conditioned on both route-level cues, such as whether the upcoming segment contains a crossing, and visual cues, such as crowd density, pedestrian motion, traffic-light state, and vehicle movement.

If $d_t=\text{proceed}$, the system dispatches the step to the Low-Level VLA for efficient local trajectory prediction. If $d_t=\text{stop-and-wait}$, the robot remains stationary and queries the High-Level VLM again after receiving a fresh observation. Complex situations, such as pedestrian crossings, dense crowds, or visually ambiguous traffic conditions, are therefore handled within the same High-Level VLM decision rather than through a second-stage fallback.

\subsubsection{Proceeding with Low-Level VLA}

For steps allowed to proceed, the Low-Level VLA receives the current image, route instruction, local goal, and recent trajectory history. These inputs are represented in the robot's egocentric frame so that the policy can focus on local motion generation.

The Low-Level VLA predicts a short-horizon trajectory:
\begin{equation}
    \hat{W}_t^{body}
    =
    \{\hat{w}_{t,1}^{body},\ldots,\hat{w}_{t,H}^{body}\}.
\end{equation}

This trajectory represents a short socially compliant motion plan toward the lookahead goal. At execution time, the predicted trajectory is converted into a near-term tracking target for the robot controller, while the full trajectory provides local motion guidance and visualization. The trajectory is re-planned at every control step using the latest observation and state estimate.

This design allows the Low-Level VLA to focus on local socially aware navigation, while the high-level route provides long-horizon guidance.

\subsubsection{Safety-Aware Stop-and-Wait Execution}

The same joint High-Level VLM output also provides the safety decision used during execution. The VLM analyzes traffic-light state, moving vehicles, pedestrian flow, crowd density, and other visually observable safety cues, and returns the decision $d_t$, confidence $conf$, and explanation $reason$ together with the routing label $r_t$.

The robot proceeds through the crossing or crowded region only when
\begin{equation}
    d_t = \text{proceed}
    \quad \text{and} \quad 
    conf \geq \alpha,
\end{equation}
If the scene is judged unsafe or the confidence is below threshold, the robot executes a stop-and-wait behavior. It remains in place, captures a fresh image after a short interval, and queries High-Level VLM again.

This process is repeated within a limited safety budget. If no confident proceed decision is reached, the episode is aborted and handed back to the operator. Once the VLM returns a confident proceed decision, the Low-Level VLA resumes local trajectory generation.

By using the High-Level VLM to provide a joint routing and safety decision at every step, \textit{Walk with Me} separates routine local control from safety-critical semantic reasoning without introducing a second-stage fallback module. This prevents crossings and crowded areas from being treated as ordinary geometric obstacles and enables explicit stop-and-wait behavior when needed.

\subsection{Closed-loop Robot Execution}

The final stage converts the selected policy output into executable robot actions and closes the perception--reasoning--action loop.

For proceed steps, the VLA prediction is combined with the current SLAM pose estimate to produce an executable low-level control command that advances the robot toward the current lookahead goal while keeping its heading aligned with the intended direction of travel. This fusion of learned trajectory prediction and SLAM feedback improves robustness to local unexpected events, such as pedestrians suddenly entering the robot's path.

For safety-critical stop-and-wait steps, no VLA trajectory is produced. In this case, the action degenerates into a zero-displacement stop action.

Before translation, the robot performs a short in-place yaw alignment toward the intended direction of motion. This keeps the forward-facing camera consistent with the route direction. The robot then tracks the target pose using its onboard SLAM feedback and built-in low-level motion controller.

After execution, the onboard localization module provides an updated pose $s_{t+1}$. The trajectory history is updated to $H_{t+1}$, the lookahead waypoint is recomputed as $w_{t+1}^{goal}$, and a new observation $o_{t+1}$ is constructed. The system then repeats the routing, policy selection, and execution process.

At each step, the system checks whether the robot has reached the destination by measuring its distance to the final waypoint. The episode terminates successfully when this distance is below an arrival tolerance. A hard iteration budget is also enforced as a safety fallback.

Overall, \textit{Walk with Me} realizes a map-free, long-horizon outdoor social navigation system by combining high-level semantic planning, adaptive routing between routine and safety-critical behaviors, and closed-loop robot execution. The semantic planner grounds abstract human intent into a concrete route, the dual-policy navigation module handles both ordinary walking and complex social situations, and the execution loop continuously updates the robot state until the destination is reached.
\section{Experiments}

We evaluate \textit{Walk with Me} in real-world outdoor environments. 
The main experiment focuses on application-level long-horizon social navigation from high-level human instructions. 
As motivated in Sec.~1, we study two representative human-centric applications: last-mile delivery and blind guidance. 
Each application contains two concrete scenarios, and each scenario is evaluated with five independent trials, resulting in 20 real-world trials in total.

Our task differs from conventional point-goal navigation, short-horizon social navigation, and route-following visual navigation. 
The robot must infer a concrete destination from an abstract human instruction, retrieve nearby POI candidates and walking routes from a public map service, execute a long-horizon outdoor route, and reason about safety in socially sensitive scenes. 
Since existing methods do not directly provide this full input-output interface, the main experiment evaluates only the complete \textit{Walk with Me} system. 
We study the influence of alternative Low-Level VLA backbones and High-Level VLM choices in the ablation studies.

\subsection{Implementation Details}

All real-world experiments are conducted on an Athena 2.0 Pro AGV wheeled robot. 
The robot is equipped with a forward-facing Intel RealSense D455 RGB-D camera, a GPS receiver, and an onboard SLAM module for estimating the local robot pose. 
The AGV chassis provides the low-level motion controller and executes target poses generated by our navigation system. 
At each control step, the robot collects the current visual observation, SLAM pose, recent trajectory history, and the current lookahead waypoint, and then sends these inputs to a remote inference server.

The Low-Level VLA is deployed on a remote server equipped with NVIDIA H20 GPUs. 
The server receives the robot observation and route context, performs VLA inference, and returns a short-horizon local trajectory, which is converted into an executable low-level control command for the AGV chassis. 
For the High-Level VLM, closed-source models are queried through their official APIs, while open-source models are deployed on the same H20-based remote inference server. 
This deployment strategy allows us to evaluate both proprietary and open-source VLM/VLA models in real-world outdoor navigation without being constrained by the onboard computation capacity of the robot.

\subsection{Main Experiment}

\paragraph{Task design}
We design two application-level task categories that reflect practical human-centric outdoor assistance scenarios: last-mile delivery and blind guidance. 
The last-mile delivery category includes delivering milk tea to Building B and package delivery to Building A, using the instructions ``Take the milk tea to Building B'' and ``Please deliver this package to Building A,'' respectively. 
The blind guidance category includes accompanying a visually impaired user for a walk and guiding the user to go shopping, with the instructions ``I want to go for a walk'' and ``I want to go shopping.'' 
Each scenario is evaluated with five independent trials, resulting in $2 \times 2 \times 5 = 20$ real-world trials in total.

Each trial starts from a high-level natural-language instruction rather than a manually specified coordinate or point goal. 
The robot must first ground the instruction into a concrete destination, construct a walking route using the public map service, and then execute the route in a closed loop.

In the last-mile delivery task, the robot receives a high-level instruction specifying the delivered item and the target building, grounds the destination, and follows pedestrian-accessible routes. 
This setting emphasizes explicit delivery instruction grounding, long-horizon route following, detour handling, and socially compliant motion in everyday outdoor environments.

In the blind guidance task, the robot guides the user toward a semantically appropriate destination while behaving conservatively in safety-critical situations. 
The two scenarios correspond to the user intents of going for a walk and going shopping. 
This task places stronger emphasis on safety reasoning and stop-and-wait behavior in socially sensitive scenes.

\begin{table}[t]
    \centering
    \caption{
    Main real-world experiment results of the complete \textit{Walk with Me} system. 
    Each application category contains two concrete scenarios, and each scenario is evaluated with five independent trials. 
    SR denotes success rate.
    }
    \label{tab:main_results}
    \resizebox{\linewidth}{!}{
    \begin{tabular}{llccc}
        \toprule
        Task
        & Scenario
        & Success / Trials
        & SR
        & Main challenge \\
        \midrule
        Last-mile Delivery
        & Milk tea to Building B
        & 4 / 5
        & 80\%
        & Intent grounding and POI selection \\
        Last-mile Delivery
        & Package to Building A
        & 3 / 5
        & 60\%
        & Long-horizon route execution \\
        \midrule
        Blind Guidance
        & Go for a walk
        & 3 / 5
        & 60\%
        & Crossing safety reasoning \\
        Blind Guidance
        & Go shopping
        & 2 / 5
        & 40\%
        & Crowd and proximity handling \\
        \midrule
        Average
        & All
        & 12 / 20
        & 60\%
        & -- \\
        \bottomrule
    \end{tabular}
    }
\end{table}

\paragraph{Evaluation metric}
We use success rate (SR) as the primary metric for the main experiment. 
A trial is considered successful if the robot reaches the selected destination within the arrival tolerance without operator takeover, safety abort, or failure to follow the generated route.

Formally, for $N$ trials, success rate is defined as
\begin{equation}
    \mathrm{SR}
    =
    \frac{1}{N}
    \sum_{i=1}^{N}
    \mathbb{I}[\text{trial } i \text{ succeeds}].
\end{equation}

We report SR for each concrete scenario, each application category, and the average over all 20 trials.

\paragraph{Results}
Table~\ref{tab:main_results} reports the main real-world evaluation results of the complete \textit{Walk with Me} system. 
Each application category contains two concrete task scenarios, and each scenario is evaluated with five independent trials. 
We report both the number of successful trials and the corresponding success rate.

The main experiment evaluates whether \textit{Walk with Me} can solve complete application-level outdoor assistance tasks rather than isolated local navigation segments. 
The two application categories cover complementary aspects of the proposed framework: last-mile delivery emphasizes explicit instruction grounding, building-level destination selection, and long-horizon route execution, while blind guidance emphasizes safety-aware behavior in socially sensitive scenes. 
Overall, \textit{Walk with Me} succeeds in 12 out of 20 real-world trials, achieving an average SR of 60\%. 
The last-mile delivery category achieves 70\% task-level SR over 10 trials, while blind guidance achieves 50\% over 10 trials. 
Within last-mile delivery, delivering milk tea to Building B is more reliable than delivering a package to Building A, suggesting that route layout and destination context can significantly affect long-horizon execution even when the task instruction is explicit. 
Blind guidance remains more challenging, especially for the shopping intent, because the robot must ground a more open-ended user request and behave conservatively in safety-critical social scenes.

\begin{table}[t]
    \centering
    \caption{
    Ablation studies on the two last-mile delivery scenarios used in the main experiment. 
    Left: High-Level VLM ablation with the Low-Level VLA fixed. 
    Right: Low-Level VLA ablation with the High-Level VLM and route-generation module fixed. 
    SR denotes success rate over 10 trials. 
    $^\ast$ denotes models deployed from our reproduced implementations.
    }
    \label{tab:ablation_modules}
    \begin{minipage}[t]{0.52\linewidth}
        \centering
        \textbf{High-Level VLM}\\[2pt]
        \resizebox{\linewidth}{!}{
        \begin{tabular}{llcc}
            \toprule
            Model
            & Type
            & Success / Trials
            & SR \\
            \midrule
            Claude-Sonnet~\cite{anthropic2025claudesonnet45}
            & Closed-source
            & 5 / 10
            & 50\% \\
            Gemini-3-Flash~\cite{google2025gemini3flash}
            & Closed-source
            & 4 / 10
            & 40\% \\
            GPT-5~\cite{openai2025gpt5systemcard}
            & Closed-source
            & 5 / 10
            & 50\% \\
            \midrule
            Qwen3-VL-8B~\cite{bai2025qwen3}
            & Open-source
            & 3 / 10
            & 30\% \\
            RoboBrain 2.0~\cite{team2025robobrain}
            & Open-source
            & 5 / 10
            & 50\% \\
            \rowcolor{orange!8} MiMo-Embodied~\cite{hao2025mimo}
            & Open-source
            & 6 / 10
            & 60\% \\
            \bottomrule
        \end{tabular}
        }
    \end{minipage}
    \hfill
    \begin{minipage}[t]{0.43\linewidth}
        \centering
        \textbf{Low-Level VLA}\\[2pt]
        \resizebox{\linewidth}{!}{
        \begin{tabular}{lcc}
            \toprule
            Method
            & Success / Trials
            & SR \\
            \midrule
            GNM$^\ast$~\cite{shah2023gnm}
            & 2 / 10
            & 20\% \\
            ViNT$^\ast$~\cite{shah2023vint}
            & 2 / 10
            & 20\% \\
            NoMaD$^\ast$~\cite{sridhar2024nomad}
            & 3 / 10
            & 30\% \\
            CityWalker$^\ast$~\cite{liu2025citywalker}
            & 5 / 10
            & 50\% \\
            \rowcolor{orange!8} SocialNav$^\ast$~\cite{chen2025socialnav}
            & 6 / 10
            & 60\% \\
            \bottomrule
        \end{tabular}
        }
    \end{minipage}
\end{table}

\subsection{Ablation Study}

We conduct ablation studies on the same two last-mile delivery scenarios used in the main experiment, \textit{\ie}, taking milk tea to Building B and delivering a package to Building A. 
These two scenarios are used because they require both high-level destination grounding and long-horizon low-level navigation, making them suitable for isolating the contributions of the High-Level VLM and Low-Level VLA modules.

\paragraph{Effect of High-Level VLM}
We first replace the High-Level VLM while keeping the Low-Level VLA and the rest of the system unchanged. 
The compared models include both closed-source and open-source VLMs. 
For closed-source models, we evaluate Claude-Sonnet~\cite{anthropic2025claudesonnet45}, Gemini~\cite{google2025gemini3flash}, and GPT~\cite{openai2025gpt5systemcard}. 
For open-source models, we evaluate Qwen3-VL-8B~\cite{bai2025qwen3}, MiMo-Embodied~\cite{hao2025mimo}, RoboBrain 2.0~\cite{team2025robobrain}, and related embodied VLMs. 
This ablation evaluates whether different VLM backbones can correctly interpret the user instruction, select the intended building-level destination from map candidates, and support successful completion of the two last-mile delivery scenarios.
As shown in Table~\ref{tab:ablation_modules}, MiMo-Embodied~\cite{hao2025mimo} obtains the best SR among the evaluated VLMs, while RoboBrain 2.0~\cite{team2025robobrain} and the strongest closed-source models achieve comparable performance. 
Qwen3-VL-8B~\cite{bai2025qwen3} achieves a lower SR in this setting, suggesting that strong general visual-language capability may still require additional adaptation for robust building-level grounding and navigation-oriented reasoning.

\begin{figure*}[t]
    \centering
    \includegraphics[width=0.88\textwidth]{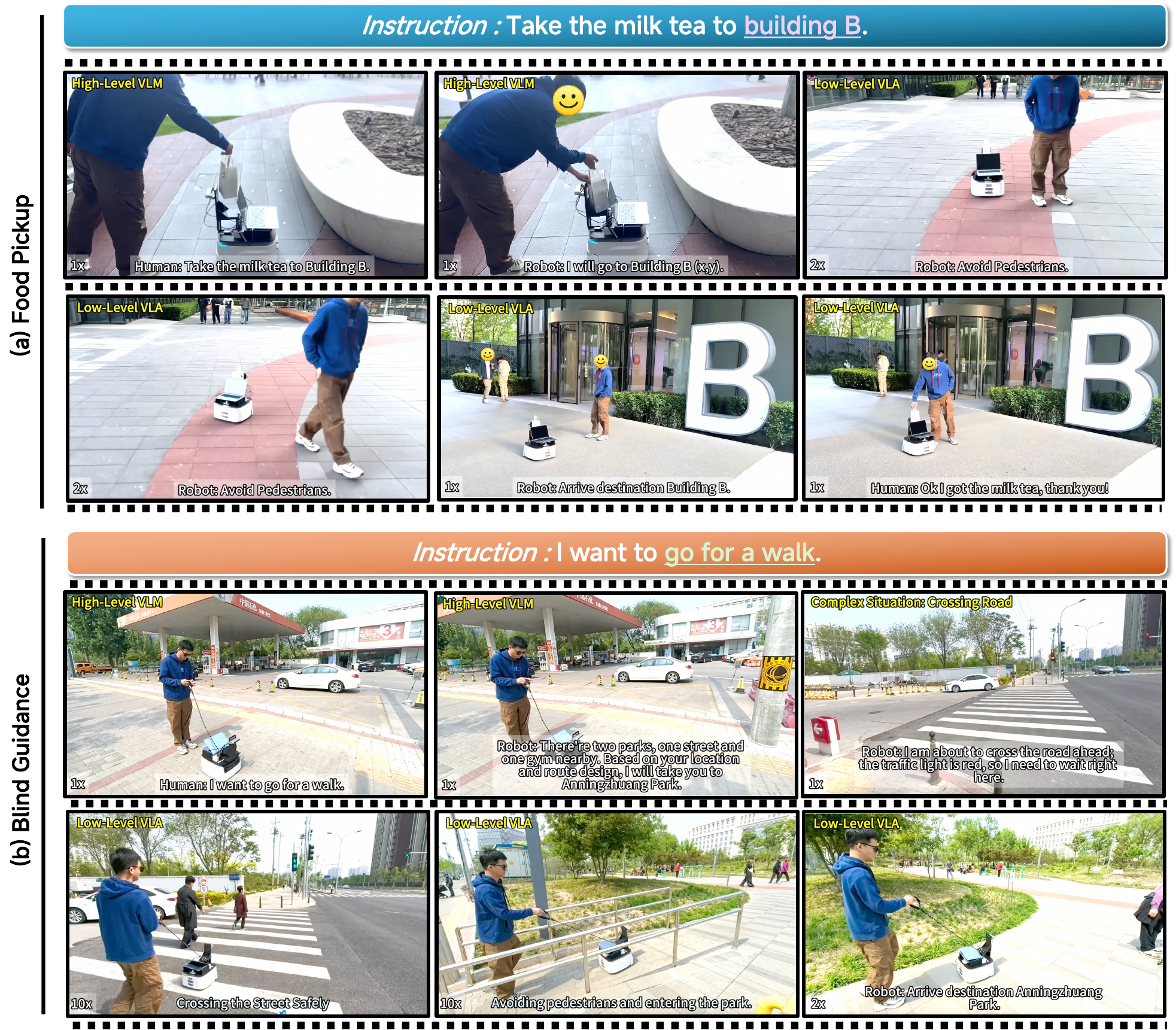}
    \caption{
    Qualitative visualization of \textit{Walk with Me} in two real-world tasks. 
    (a) In last-mile delivery, the system grounds the instruction ``Take the milk tea to Building B,'' navigates toward Building B, avoids nearby pedestrians, and completes the delivery. 
    (b) In blind guidance, the system grounds ``I want to go for a walk'' to a nearby park, reasons about a road-crossing scene, and continues toward the destination while avoiding pedestrians.
    }
    \label{fig:qualitative}
\end{figure*}

\paragraph{Effect of Low-Level VLA}
We then replace the Low-Level VLA while keeping the High-Level VLM and global route generation unchanged. 
This setting evaluates whether different low-level navigation policies can follow the same route context and handle social navigation events encountered during last-mile delivery. 
Table~\ref{tab:ablation_modules} reports the success rate of each High-Level VLM and Low-Level VLA variant on the same two last-mile delivery scenarios.
The Low-Level VLA ablation shows a clearer performance gap across navigation backbones. 
Classical goal-conditioned visual navigation models such as GNM~\cite{shah2023gnm} and ViNT~\cite{shah2023vint} achieve lower SR, while NoMaD~\cite{sridhar2024nomad} and CityWalker~\cite{liu2025citywalker} improve performance by producing more robust local navigation behavior. 
SocialNav~\cite{chen2025socialnav} achieves the highest SR, indicating that a low-level policy trained or designed for socially aware navigation better handles turns, detours, pedestrian proximity, and other local interactions encountered during real-world last-mile delivery.

\begin{figure*}[t]
    \centering
    \includegraphics[width=0.88\textwidth]{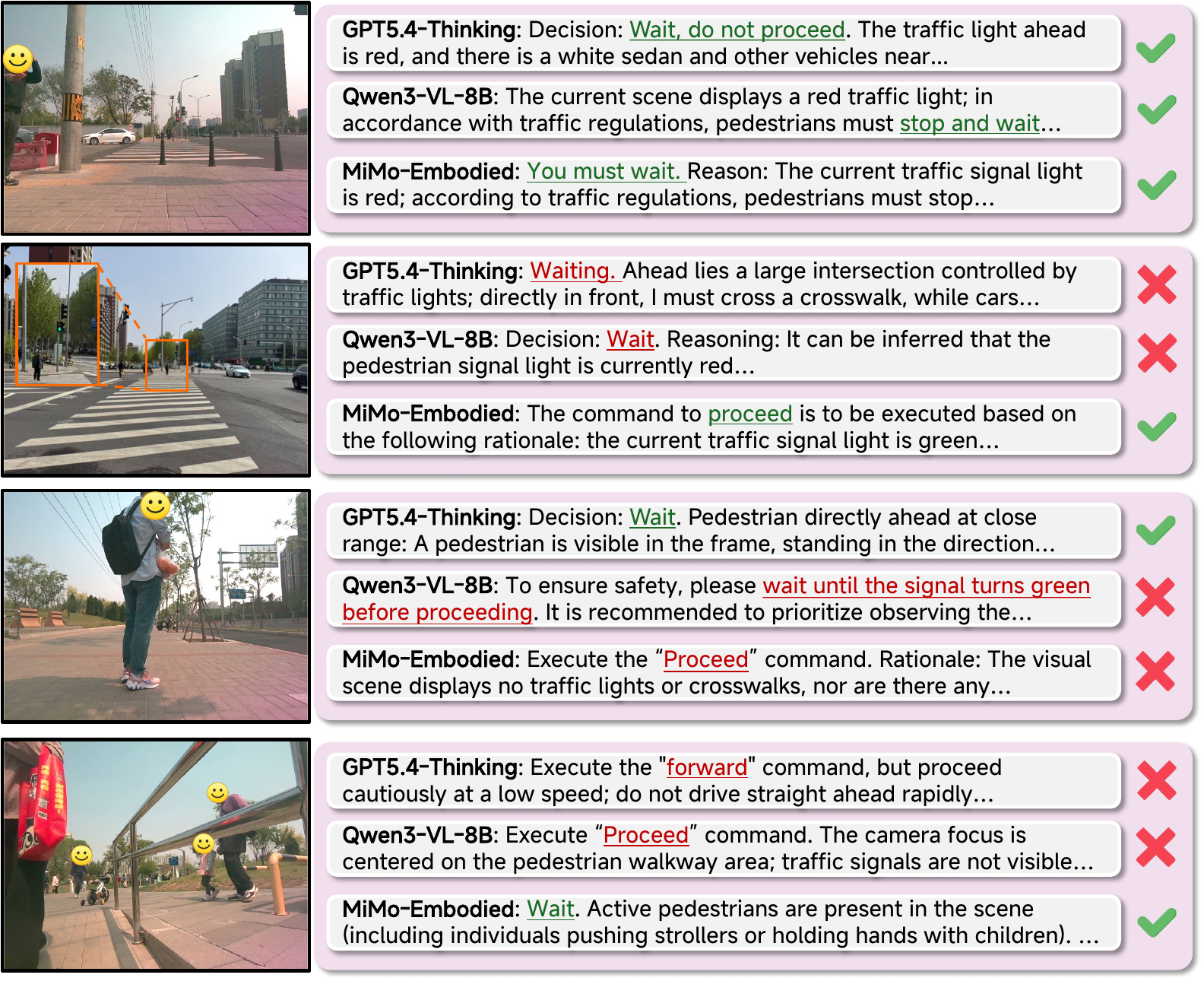}
    \caption{
    Qualitative visualization of High-Level VLM reasoning in safety-critical scenes.
    The High-Level VLM jointly assesses the scene complexity and stop-and-wait decision from visual observations and route context. The examples compare model responses across crossing, pedestrian-proximity, and crowded-scene situations.
    }
    \label{fig:qualitative_vlm}
\end{figure*}

\begin{figure*}[t]
    \centering
    \includegraphics[width=0.88\textwidth]{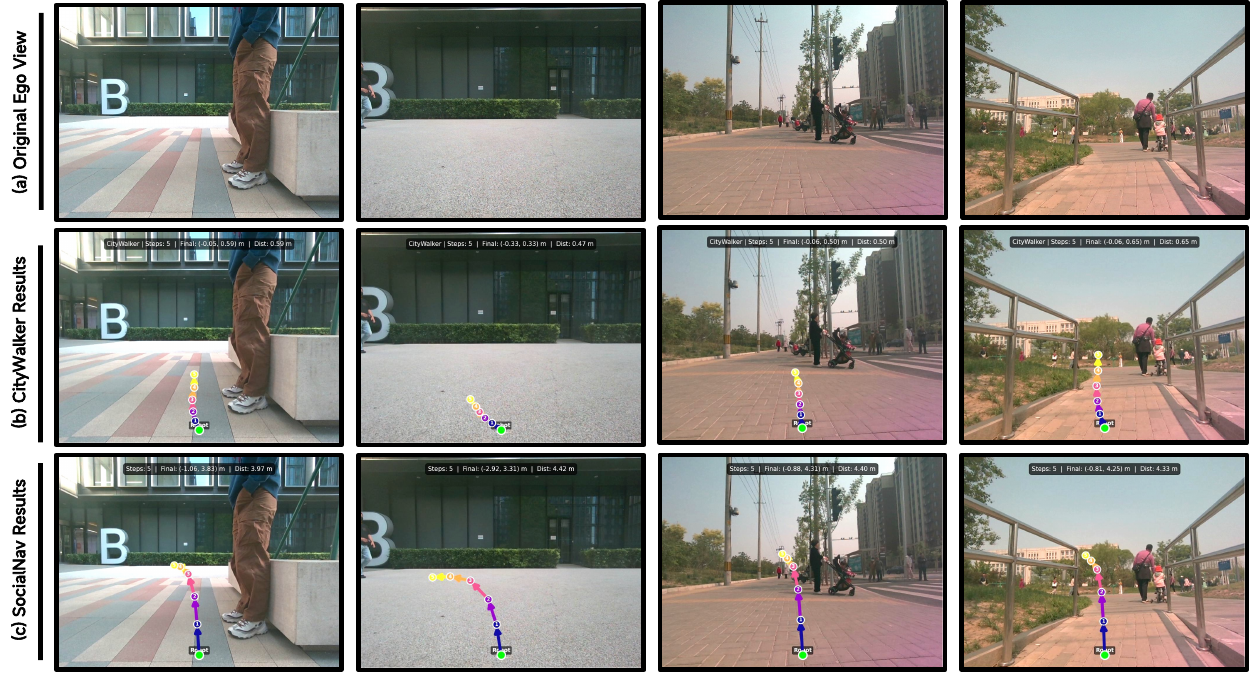}
    \caption{
    Qualitative visualization of Low-Level VLA trajectory prediction.
    Given the current visual observation, route instruction, local goal, and recent trajectory context, the Low-Level VLA predicts a short-horizon socially compliant trajectory that guides the robot toward the next waypoint while accounting for nearby pedestrians and local scene geometry.
    }
    \label{fig:qualitative_vla}
\end{figure*}

\subsection{Qualitative Analysis}

Fig.~\ref{fig:qualitative} shows two representative real-world executions. 
In the last-mile delivery case, the High-Level VLM interprets the user request and confirms Building B as the delivery destination, while the Low-Level VLA handles local motion around pedestrians before the robot reaches the building entrance. 
In the blind guidance case, the system maps the open-ended walking request to a nearby park and uses high-level safety reasoning at the road crossing, stopping when the traffic light is unsafe and resuming only after the crossing becomes safe. 
After crossing, the Low-Level VLA continues route-conditioned navigation while avoiding pedestrians along the path. 
These examples illustrate how \textit{Walk with Me} combines instruction grounding, explicit safety reasoning, and socially aware local control in a single closed-loop outdoor navigation process.

Fig.~\ref{fig:qualitative_vlm} visualizes High-Level VLM reasoning in representative safety-critical scenes. The models generally capture whether the robot should proceed or wait based on crossing context, pedestrian flow, proximity, and other visual cues. At the same time, their responses can diverge under ambiguous outdoor observations, suggesting that High-Level VLM reasoning for fine-grained safety assessment still requires further optimization.

Fig.~\ref{fig:qualitative_vla} further visualizes the Low-Level VLA trajectory prediction during local execution. The predicted trajectories provide near-term motion guidance conditioned on the route instruction and current observation, while the robot executes them together with SLAM-based state feedback and the onboard low-level controller. This allows the system to follow the long-horizon route while reacting to local changes, such as pedestrians entering the robot's path.
\section{Conclusion}

In this work, we presented \textbf{Walk with Me}, a map-free framework for long-horizon outdoor social navigation from high-level human instructions. Instead of relying on pre-built HD maps or manually specified point goals, the proposed system grounds abstract user intent with public map-service POIs, constructs coarse geo-referenced waypoints, and executes the route through a closed-loop hierarchy that combines high-level VLM reasoning with low-level VLA control. The observation-aware router further enables the robot to distinguish routine navigation from safety-critical situations, invoking explicit VLM-based reasoning and stop-and-wait behavior at crossings, crowds, and ambiguous outdoor scenes. Real-world experiments on a commodity wheeled robot show that \textit{Walk with Me} can complete application-level last-mile delivery and blind guidance tasks across long-horizon outdoor routes, while the ablation studies highlight the importance of both navigation-oriented high-level reasoning and socially aware low-level action generation. These results suggest that combining lightweight geographic priors, semantic reasoning, and learned local control is a promising direction for practical human-centric robot assistance. At the same time, the current system remains limited by the reliability of public map services, GPS and localization noise, and the robustness of VLM/VLA models in rare or highly dynamic social scenes. Future work will explore stronger uncertainty handling, richer multi-modal perception, and broader deployment across more diverse outdoor environments.


\bibliographystyle{plainnat}
\bibliography{main}


\end{document}